\documentclass[sigconf,authorversion,nonacm]{acmart} 

\AtBeginDocument{%
  \providecommand\BibTeX{{%
    \normalfont B\kern-0.5em{\scshape i\kern-0.25em b}\kern-0.8em\TeX}}}

\usepackage{array}
\usepackage{graphicx}
\usepackage{multirow}
\usepackage{longtable}
\usepackage{pdflscape}
\usepackage{float}
\usepackage{dblfloatfix}

\newcolumntype{L}[1]{>{\raggedright\let\newline\\\arraybackslash\hspace{0pt}}m{#1}}
\newcolumntype{C}[1]{>{\centering\let\newline\\\arraybackslash\hspace{0pt}}m{#1}}
\newcolumntype{R}[1]{>{\raggedleft\let\newline\\\arraybackslash\hspace{0pt}}m{#1}}
\copyrightyear{2024} 
\acmYear{2024} 
\setcopyright{rightsretained} 
\acmConference[FAccT '24]{The 2024 ACM Conference on Fairness, Accountability, and Transparency}{June 3--6, 2024}{Rio de Janeiro, Brazil}
\acmBooktitle{The 2024 ACM Conference on Fairness, Accountability, and Transparency (FAccT '24), June 3--6, 2024, Rio de Janeiro, Brazil}\acmDOI{10.1145/3630106.3658911}
\acmISBN{979-8-4007-0450-5/24/06}

\begin{document}

\title[Not My Voice! A Taxonomy of Ethical and Safety Harms of Speech Generators]{Not My Voice! A Taxonomy of Ethical and Safety Harms of Speech Generators}

\author{Wiebke Hutiri}
\email{wiebke.hutiri@sony.com}
\affiliation{%
  \institution{Sony AI}
  \country{Switzerland}
}
\orcid{}
\author{Orestis Papakyriakopoulos}
\email{orestis.p@tum.de}
\affiliation{%
  \institution{Sony AI}
  \country{Switzerland}
}
\affiliation{%
  \institution{Technical University of Munich}
  \country{Germany}
}
\author{Alice Xiang}
\email{alice.xiang@sony.com}
\affiliation{%
  \institution{Sony AI}
  \country{USA}
}

\begin{abstract}

The rapid and wide-scale adoption of AI to generate human speech poses a range of significant ethical and safety risks to society that need to be addressed. For example, a growing number of speech generation incidents are associated with swatting attacks in the United States, where anonymous perpetrators create synthetic voices that call police officers to close down schools and hospitals, or to violently gain access to innocent citizens' homes. Incidents like this demonstrate that multimodal generative AI risks and harms do not exist in isolation, but arise from the interactions of multiple stakeholders and technical AI systems. In this paper we analyse speech generation incidents to study how patterns of specific harms arise. We find that specific harms can be categorised according to the exposure of affected individuals, that is to say whether they are a subject of, interact with, suffer due to, or are excluded from speech generation systems. Similarly, specific harms are also a consequence of the motives of the creators and deployers of the systems. Based on these insights we propose a conceptual framework for modelling pathways to ethical and safety harms of AI, which we use to develop a taxonomy of harms of speech generators. Our relational approach captures the complexity of risks and harms in sociotechnical AI systems, and yields a taxonomy that can support appropriate policy interventions and decision making for the responsible development and release of speech generation models.
\end{abstract}


\maketitle

\section{Introduction}
\label{s:introduction}

The wide-scale deployment and adoption of generative AI systems poses a range of significant risks to society, some of which have already resulted in material harm to people and communities. Particularly, speech generation, colloquially referred to as audio deepfakes, is fuelling an unprecedented wave of malicious activity~\cite{ftc2023}. However, evaluating risks and targeting interventions to mitigate harms of speech generation technology (or speech generators) remains challenging, as these complex, multimodal generative AI systems interact with human stakeholders in diverse and unexpected ways. For example, in a growing number of swatting attacks in the United States, anonymous perpetrators create synthetic voices that call police officers to close down schools and hospitals to investigate bogus bomb threats, or to violently gain access to innocent citizens' homes~\cite{cox2023aamerica}. The same type of speech generator was used in a different incident to synthesise voices of deceased children so that video clones could vocalise narratives of their brutal murders on a viral social media channel~\cite{hassan2023aifor}. 

Incidents like these demonstrate that risks and harms of speech generators are not merely features or failures of the technology. Rather, speech generators exist within complex sociotechnical systems, whose (mis)behaviour arises from interactions between networks of human actors and technical subsystems~\cite{deBruijn2009SystemSystems}. To model and intervene in the behaviour of complex sociotechnical systems, it is necessary to understand not only the technical subsystems, their dependencies and objectives, but also the roles, interests and relationships between strategically acting, independent actors. Reflecting this perspective back to AI risks and harms, this paper positions that interventions aimed at mitigating risks of generative AI require an understanding of \textit{what risks exist \underline{and} how risks arise} from interactions of stakeholders with generative AI. 

To understand risks of AI systems, recent studies have developed taxonomies of risks and harms of AI~\cite{OECD2023StocktakingDefinition, Hoffmann2023CSETGuide, Shelby2022SociotechnicalReduction, Weidinger2022TaxonomyModels, Lee2023DeepfakesRisks, Critch2023TASRA:AI}. While these taxonomies present helpful categorisation schemes for classifying what \textit{specific harm} was inflicted in an incident, they do not consider who was harmed (i.e. the \textit{affected entity}), who is accountable for inflicting the harm (i.e. the \textit{responsible entity}), and how this impacts which specific harms materialised. In speech generation incidents, the \textit{affected} and \textit{responsible entities} are not always obvious. Moreover, several specific harms often co-occur. For example in the swatting incidents described previously, the responsible entities are anonymous perpetrators who deploy speech generators to harm, but also police officers who are compelled to take violent action in response to their interaction with speech generators. This obscures whether police officers who are deceived by speech generators are a responsible or an affected entity. Citizens that are victims of the swatting attacks do not interact with speech generators at all, but may suffer bodily harm and psychological distress, or harm to their property. Furthermore, the swatting incidents also pose a national security threat, threat of terrorism and abuse of public services that may harm US citizens at large. 

Even though it is clear that speech generators can lead to severe harms at large scale, existing taxonomies only provide limited support for modelling the interactions between harms and stakeholders, and for categorising new specific harms of generative speech technologies. Given the difficulties of adopting existing taxonomies to speech generators, the aim of this paper is thus two-fold. First, we propose a conceptual framework for modelling pathways to ethical and safety harms of generative AI systems. We adopt the idea of \textit{pathways to harm} from environmental risk assessments, where pathways capture a ``causal chain of events required for a harm to be realised''~\cite[p. 6]{Connolly2022RecommendationsControl}. In our context we consider pathways as the interactions between affected entities, responsible entities, speech generators and specific harms. We then use the conceptual framework to develop a taxonomy of harms of speech generators. Our study uses a design science research approach, where we develop the conceptual model and taxonomy iteratively and in parallel, drawing on existing taxonomies and reported incidents of harm from speech generators. 



\section{Related Work}
\label{s:related_work}

A \textit{harm} is a negative outcome that entails value loss or damage to people, a \textit{hazard} a source or situation with potential for harm and a \textit{risk} the likelihood and severity of harm when exposed to a hazard~\cite{ISO24748}. This section reviews AI risk and harm taxonomies from literature and incident databases, and research on risks and harms of speech generators.

\subsection{AI Incident Databases and Annotation Taxonomies}
The OECD AI Incidents Monitor~\cite{oecdai_aim}, the AI Incident Database (AIID)~\cite{McGregor2021PreventingDatabase, raic_aiid}, and the AI, Algorithmic, and Automation Incidents and Controversies (AIAAIC) database~\cite{aiaaic} actively maintain records of reported AI hazards and harms. Each of these databases has developed a taxonomy to categorise harms in order to support the annotation of AI incidents.

In the OECD AI Incident database, harms are the starting point for defining an AI incident. Harms are categorised into harm type, level of severity, scope and geographic scale~\cite{OECD2023StocktakingDefinition}. Furthermore, harms can be tangible or intangible, quantifiable or unquantifiable, potential (i.e. not materialised) or actual (i.e. materialised), reversible or irreversible, once-off or recurring with cumulative effects, and have direct or indirect impact. The CSET taxonomy~\cite{Hoffmann2023CSETGuide} has been developed to annotate incidents in the AIID. It includes various harm types, such as physical health or safety, financial loss, infrastructure, natural environment, violation of human rights, and other tangible and intangible harms. The CSET taxonomy allows for an incident to have multiple harms and second occurrences of harm. It also states that harm designations can change over time. The AIAAIC classifies its database entries as system, incident, issue and data. Only incidents are harms or hazards. The AIAAIC defines harms as actual negative external or internal impacts of an incident, system or dataset. External impacts are individual, societal or environmental, while internal impacts can be strategic or reputational, operational, financial, and legal or regulatory. In addition to harm types, the AIAAIC identifies 15 risk categories, such as accuracy or reliability, anthropomorphism, bias or discrimination, surveillance, and transparency.  

\subsection{Taxonomies of AI Risks and Harms}

\citet{Shelby2022SociotechnicalReduction} derive a framework for classifying harms in algorithmic system from literature using a scoping review. They categorise harms into 5 themes. Algorithmic systems lead to \textit{representational harms} when they reinforce subordination of social groups. \textit{Allocative harms} arise from opportunity, resources or information being withheld from marginalised groups. Systems that fail disproportionately for certain groups lead to \textit{quality-of-service harms}. \textit{Interpersonal harms} affect relationships between people and communities, and may also affect individuals themselves. Finally, \textit{social system harms} adversely affect society at large. In addition to identifying these overarching themes, the authors categorise each theme into sub-types of harms, and list specific harms for each sub-type. \citet{Weidinger2022TaxonomyModels} developed a taxonomy of risks posed by large language models from discussions and workshops with experts and a literature review. Many of the types of harm presented in this taxonomy are also captured as specific harms by \citet{Shelby2022SociotechnicalReduction}. Weidinger et al. further revise their taxonomy in~\cite{Weidinger2023SociotechnicalSystems}. This taxonomy captures six high-level risk areas: \textit{representation and toxicity harms}, \textit{misinformation harms}, \textit{information and safety harms}, \textit{malicious use}, \textit{human autonomy and integrity harms} and \textit{socio-economic and environmental harms}. Building on these taxonomies, \citet{Kirk2023PersonalisationFeedback} develop a taxonomy of risks that are specific to large language models which are personalised based on user feedback. New risks that they identify are user \textit{effort} and volunteer labour required for providing feedback, \textit{essentialism and profiling} leading to non-consensual categorisation of people, and an increased tendency to \textit{anthropomorphism}. 

Several studies have used incident databases for creating purpose-specific taxonomies and for structuring failure analysis of AI systems. Leet et al.'s~\cite{Lee2023DeepfakesRisks} taxonomy of AI privacy risks codifies patterns from privacy incidents catalogued in the AIAAIC that arise as a result of AI capabilities and data requirements. The authors systematically reviewed incidents to identity privacy harms and assess whether AI creates or exacerbates the potential for harm. They coded incidents into four high-level risk categories and associated harms that have been established in Solove's privacy taxonomy~\cite{solove2006APrivacy}: \textit{data collection risks}, \textit{data processing risks}, \textit{data dissemination risks} and \textit{invasion risks}. 

\citet{Pittaras2023AIncidents} developed a taxonomy to support failure cause analysis of incidents in the AIID. Their taxonomy combines system \textit{goals}, \textit{methods and technologies} used for system implementation, and technical \textit{failure causes} that result in technical system misbehaviour. \citet{Chanda2022OmissionFailures} study how 28 factors arising during system development can lead to AI failures that result in harms. They summarise the failures into two types of errors, namely \textit{errors of commission}, or design flaws, and \textit{errors of omission}, or implementation flaws. \citet{Raji2022TheFunctionality} offer an expanded view of AI system failures beyond technical aspects, and consider how the assumption of system functionality by creators and promoters of AI influences failure. Drawing from incidents in the AIAAIC database, they present a taxonomy of AI system failures, which they categorise as conceptually and practically \textit{impossible tasks}, \textit{engineering failures}, \textit{post-deployment failures}, and \textit{communication failures}.

\subsection{Risks and Harms of Speech Generators}

Since March 2023, the OECD AI Incident database has seen a sharp increase in the number of \textit{voice} related incidents. The increase in incidents reflects the growing societal and political malaise about speech generation systems, more colloquially referred to as voice cloning. In fact, voice cloning is one of the things that keeps the US's chief AI strategist up at night~\cite{varanasi2023BidenNight}. As if in defiance of the rapid rise in speech generation incidents, the majority of researchers in the domain point out potential positive applications of their research, but rarely consider negative consequences~\cite{Barnett2023TheReview}. From a review of music and speech generation literature, \citet{Barnett2023TheReview} positions that many previously identified generative AI harms~\cite{Weidinger2022TaxonomyModels} persist in the speech generation domain. In addition, the author identified new specific harms related to the \textit{loss of agency and authorship}, \textit{copyright infringements}, \textit{overuse of speaker data}, \textit{non-consensual use of biometric data}, and \textit{deepfakes}. In AI ethics and safety research, evaluations of the hazards and harms of speech generation systems remain limited~\cite{Weidinger2023SociotechnicalSystems}, despite an increase in related research on text and image generation systems~\cite{Bender2021OnBig, Luccioni2023StableModels, Bianchi2023EasilyScale, Seshadri2023TheGeneration}. While prior work has considered attributes pertaining to human perception of speech synthesis~\cite{Cambre2019OneDevices, Cambre2020ChoiceContent, Scott2020HumanVoices}, concerns around affective speech synthesis and conversion~\cite{Triantafyllopoulos2022AnEra} and ethics of singing voice synthesis~\cite{Lee2022EthicsDevelopers}, these works focus on a narrow spectrum of risks, and do not account for the scale, scope and diversity of emerging speech generation systems.

\subsection{Shortfalls of Existing Taxonomies}
Taxonomies of risk can provide a valuable starting point for understanding and studying risks and harms of speech generators. However, existing taxonomies have several shortcomings. General taxonomies such as~\cite{Shelby2022SociotechnicalReduction, Weidinger2022TaxonomyModels, Weidinger2023SociotechnicalSystems} lack a conceptual framework that makes categorisation choices explicit. Consequently it is not easy to adapt or extend the taxonomies to new domains and comparing taxonomies is not intuitive. For example, many speech generation incidents use the technology for impersonation. However, incidents that involve impersonation are not always associated with fraud, as the taxonomy of \citet{Weidinger2023SociotechnicalSystems} suggests. Furthermore, an identity harm such as impersonation is rarely the only harm in a speech generation incident, and the person being impersonated is not the only affected entity. Instead, impersonation is frequently used as a means to achieve other ends, such as trolling, electoral sway or social polarisation. Many of these ends can be considered as \textit{malicious technology use} (as per Weidinger et al.), but this high-level categorisation is too broad to offer actionable insights. Moreover, identity related harms are so instrumental to speech generators, that they should be considered in a more nuanced way, with impersonation harms being distinguishable from identity theft, identity hijack, copyright theft and violation of right to publicity.

Beyond categorising harms, taxonomies derived from literature do not provide insights on how harms arise~\cite{Barnett2023TheReview}. Taxonomies for predictive AI (e.g. \citet{Shelby2022SociotechnicalReduction}) do not account for harms specific to generative AI, and taxonomies of risks of generative AI focus mostly on categorising harms of large language models, less on multimodal AI~\cite{Weidinger2023SociotechnicalSystems}. While taxonomies derived from incidents for specific purposes such as privacy~\cite{Lee2023DeepfakesRisks} or failure analysis~\cite{Pittaras2023AIncidents, Chanda2022OmissionFailures, Raji2022TheFunctionality} overcome these shortfalls, they remain focused on their specific area of interest. They are thus applicable to speech generation where privacy or failure analysis is concerned, but they do not capture specific harms or pathways to harm of speech generators. Thus, there remains a need for an AI harm taxonomy that captures specific harms and that supports modelling pathways to harm of speech generators.

\section{Overview of Speech Generation}
\label{s:background}

We now provide an overview of speech synthesis and speech generation systems, categorise speech generation tasks and factors that affect their performance, and summarise recent technical advances that are increasing their capabilities. 

\subsection{Introduction to Speech Synthesis}
The overall goal of a speech synthesis system is to transform text to speech, that is to convert a written character input to an acoustic waveform~\cite{itsp2022}. To do this, a basic speech synthesis system includes a text analysis module and a synthesis algorithm. The text analysis module converts written text into a linguistic representation that contains information about the pronunciation, intonation, rhythm and emphasis. The synthesis algorithm takes the linguistic representation as input, and returns an acoustic waveform as output. 

\begin{table*}[h!]
\begin{tabular}{L{0.55\linewidth}|L{0.4\linewidth}}
\textbf{primary prompt + [secondary prompt] → output} & \textbf{Task} \\ \midrule
text\textsubscript{1}\textsuperscript{A} → text\textsubscript{1}\textsuperscript{B} & Machine Translation (A→B) \\
text\textsubscript{1}\textsuperscript{A} → speech\textsubscript{1}\textsuperscript{A0} & Text-to-Speech (TTS) \\
text\textsubscript{1}\textsuperscript{A} + [text prompt]\textsubscript{2}\textsuperscript{Ax} → speech\textsubscript{1}\textsuperscript{A0x} & emotional / styled TTS \\
text\textsubscript{1}\textsuperscript{A} + [speech prompt]\textsubscript{2}\textsuperscript{Ai} → speech\textsubscript{1}\textsuperscript{Ai} & Voice Cloning (VoCl) \\
text\textsubscript{1}\textsuperscript{A} + [speech prompt]\textsubscript{2}\textsuperscript{Bi} → speech\textsubscript{1}\textsuperscript{Ai} &  Cross-lingual (XL) TTS / VoCl (B→A) \\
speech\textsubscript{1}\textsuperscript{Ai} → speech\textsubscript{1}\textsuperscript{Bi} & Speech-to-Speech (S2S) Translation (A→B) \\
speech\textsubscript{1}\textsuperscript{Ai} + [text prompt]\textsubscript{2}\textsuperscript{Ax} → speech\textsubscript{1}\textsuperscript{Aix} & Prompted Voice Conversion (VoCo)\\
speech\textsubscript{1}\textsuperscript{Ai} + [speech prompt]\textsubscript{2}\textsuperscript{Aj} → speech\textsubscript{1}\textsuperscript{Aj} & Text-free VoCo \\
speech\textsubscript{1}\textsuperscript{Aix} + [speech prompt]\textsubscript{2}\textsuperscript{Aiy} → speech\textsubscript{1}\textsuperscript{Aiy} & Emotional VoCo \\
speech\textsubscript{1}\textsuperscript{Ai} + [speech prompt]\textsubscript{2}\textsuperscript{Bj} → speech\textsubscript{1}\textsuperscript{Aj} & Cross-lingual VoCo (B→A) \\
\end{tabular} \medskip
\caption{Overview of Speech Generation Tasks. The following superscripts are used. \textbf{A, B}: different languages | \textbf{0}: default/generic voice | \textbf{i, j}: different speaker identities | \textbf{x, y}: different speaking styles, accents, intonations, emotions, etc.}
\label{tab:speech_gen_tasks}
\vspace{-1em}
\end{table*}

As in other domains, neural networks have lead to rapid progress in the field over the past decade~\cite{Tan2021ASynthesis}. The dominant text-to-speech (TTS) approaches now use neural speech synthesis to predict acoustic features from characters or phonemes using an acoustic model. Acoustic features are then vocalised with a vocoder to output a waveform. The different system architectures for doing this can be classified based on their dataflow from text input to acoustic features and waveform, their neural network structure, the type of generative model they use, and whether token prediction is sequential (autoregressive) or parallelised (non-autoregressive). Recently, end-to-end architectures that combine the acoustic model and the vocoder to output a waveform directly from characters or phonemes have been proposed~\cite{Ren2020FastSpeechSpeech, Kim2021ConditionalText-to-Speech, Casanova2021YourTTS:Everyone}. 

\subsection{Speech Generation Tasks}

TTS systems are a category of speech generation systems, which use and produce multimodal text, speech and audio inputs and outputs. The input to a speech generation systems consists of at least one prompt, which can be text or speech. The system output is always speech, but can also include other audio sounds (e.g. music or background noise). In Table~\ref{tab:speech_gen_tasks} we show an overview of the various configurations of input prompts and resultant output speech for different speech generation tasks. We introduce the following convention to describe input prompts and distinguish between tasks. The primary prompt (subscript\textsubscript{1} in Table~\ref{tab:speech_gen_tasks}) is the prompt that provides the semantic content for the output. The secondary prompt (subscript\textsubscript{2}) provides information about the speech style, accent, emotion or other attributes of the generated speech. Typically the secondary prompt is a speech utterance whose acoustic features are to be transferred to the primary prompt. However, recent advances in speech generation also accept a secondary text prompt~\cite{Ji2023TextrolSpeech:Models, Liu2023PromptStyle:Descriptions, Guo2023Promptts:Descriptions, Leng2023PrompttsPrompt}. This makes it possible to control the generated speech with written text (e.g. \textit{"When I was a young child, I used to dream of the hills."} [in the voice of an old man]). 

For TTS and voice cloning (VoCl) tasks, the primary prompt is written text. While a standard TTS system generates speech for a generic, pre-trained voice, voice cloning synthesises speech for the primary text prompt to sound like it has the identity of the speaker of the secondary prompt. For speech-to-speech (S2S) translation and voice conversion (VoCo), the primary prompt is a speech utterance. Voice conversion can transfer a variety of speech attributes, such as identity, emotion, accent and speaking style, from the secondary prompt to the primary prompt~\cite{Zhou2022EmotionalESD, Schroder2001EmotionalReview}. In contrast to S2S translation which directly converts speech from one language to another, cross-lingual (XL) voice cloning and voice conversion accept a secondary speech prompt in a different language to the primary speech prompt, and transfer the acoustic features of the voice of the secondary prompt to the text or speech in the primary prompt. The translation thus happens through indirect means.

\subsection{Advances in Speech Generation Capabilities}
Neural networks have improved the naturalness, robustness and quality of speech synthesis systems, while allowing for faster training and greater efficiency. However, they largely remain limited to synthesising speech for a specific speaker, in a specific language. Moreover, to use them for more advanced speech generation tasks like speech translation, they need to be combined in a cascade system architecture with automatic speech recognition and machine translation. Unified speech models with text and speech pre-training can overcome these constraints~\cite{Ao2022SpeechT5, Zhang2023SpeakModeling, Wang2023NeuralSynthesizers, Wang2023VioLA:Translation}. These models can take text or speech as input, synthesise speech for different speakers and for multiple languages. They expand the capabilities of single-task systems (like TTS) to include a diverse set of adjacent speech generation tasks. While the capabilities of speech generation systems are increasing, the technical barriers to entry for using them are reducing. Without intervention, this creates the preconditions for a proliferation of speech generator related harms.

\section{Research Approach}
\label{s:method}

\begin{figure}[b]
    \centering
    \includegraphics[width=\linewidth]{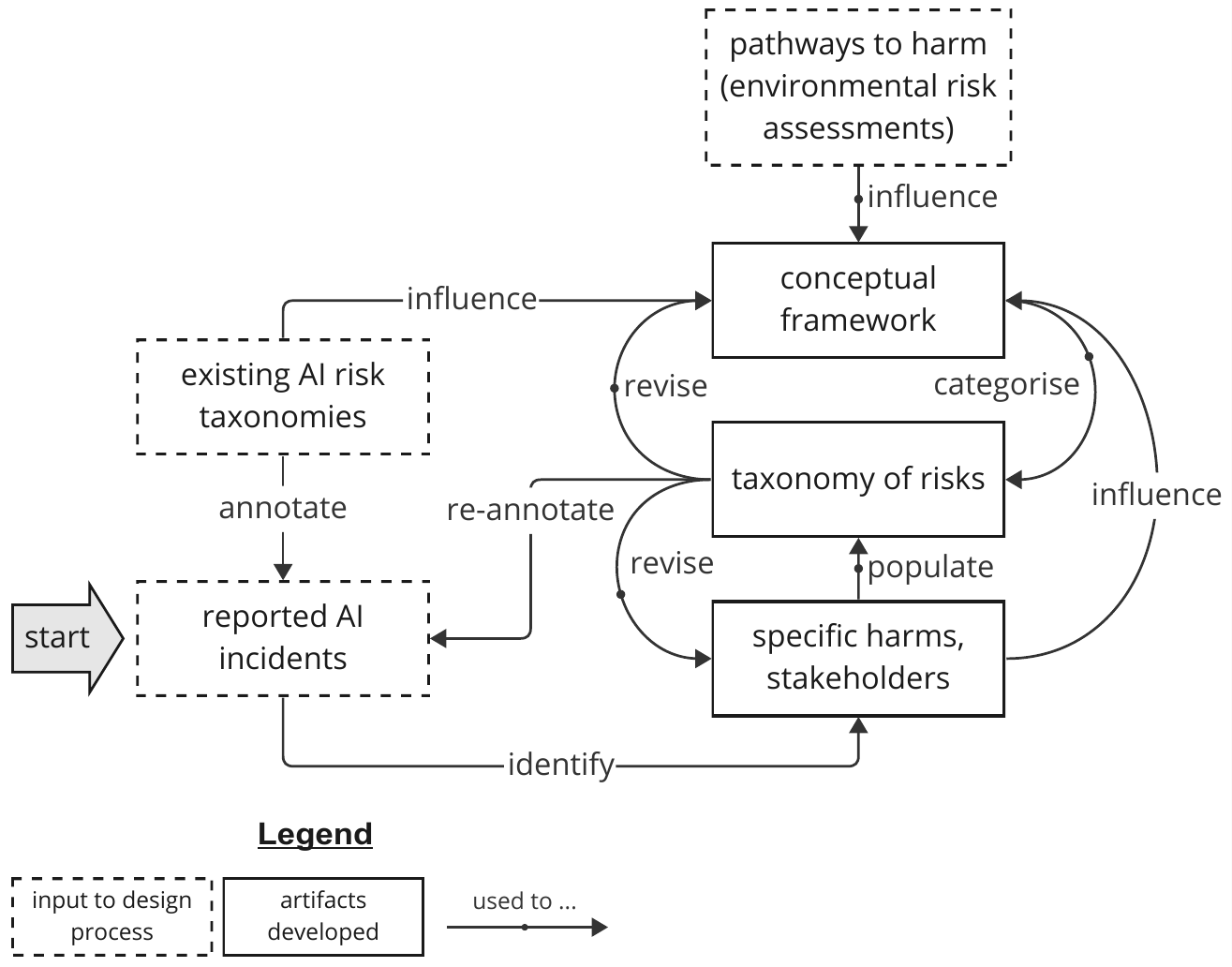}
    \caption{Iterative approach for developing conceptual framework and taxonomy of harms of speech generators}
    \label{fig:method}
\end{figure}

We used an iterative design science research approach~\cite{Gregor2013PositioningImpact} to develop the conceptual framework for modelling pathways to AI harms and the taxonomy of harms of speech generators in parallel (see Figure~\ref{fig:method}). We first identified relevant incidents that occurred between 2019 and 2023 with help of the OECD AI incident database, the AIID, the AIAAIC database and from an internal incident database. We found 35 reported AI incidents where harms could be traced back to speech generation tasks. 

\begin{figure*}[t]
    \centering
    \includegraphics[width=\textwidth]{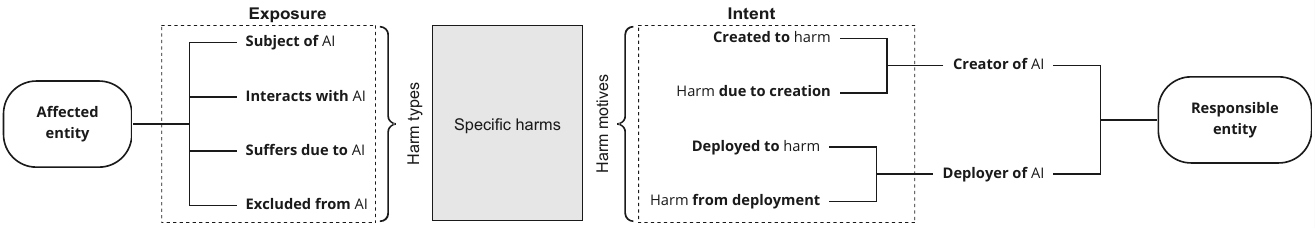}
    \vspace{-1.5em}
    \caption{Conceptual framework for modelling pathways to AI harms}
    \label{fig:ai_harm_framework}
\end{figure*}

Our point of departure for annotating the incidents were the specific harms, harm sub-types and overarching themes proposed by existing AI risk taxonomies~\cite{Shelby2022SociotechnicalReduction, Weidinger2023SociotechnicalSystems, OECD2023StocktakingDefinition, Hoffmann2023CSETGuide, Lee2023DeepfakesRisks}. We identified specific harms and the main stakeholders involved in each incident, and annotated it accordingly. The collection of specific harms and stakeholders was then used to populate a preliminary taxonomy. However, we found it difficult to harmonise the categories of existing taxonomies into a scheme that captured the salient characteristics of speech generation harms. We thus turned to theories on pathways to harm from the adjacent environmental risk assessment domain~\cite{Connolly2022RecommendationsControl} to develop a conceptual framework for modelling pathways to AI harms. The conceptual framework was influenced by the stakeholder types we observed in the annotated incidents, and by existing taxonomies which contained harm types that we did not observe in incidents, but which we thought were likely to occur in future. The conceptual framework was then applied to categorise our taxonomy of harms into harm types specific to speech generators. Several iterations followed in which we used the taxonomy of harms to re-annotate reported incidents, revised specific harms and stakeholders, and the conceptual framework. The annotated incidents of our final iteration are listed in Table~\ref{tab:speech_gen_incidents} in Appendix~\ref{a:incidents}. 

\section{Conceptual Framework}
\label{s:framework}

Conceptually, we frame pathways to AI harms as follows. AI harms are caused by responsible entities that create or deploy AI, and result in negative outcomes to affected entities. On the one hand AI harms are a consequence of an affected entity's \textit{exposure} to AI. This exposure can be of different kinds: harm can arise when an affected entity is the \textit{subject of}, \textit{interacts with}, \textit{suffers due to}, or is \textit{excluded from} AI. An AI incident may involve several affected entities who are subjected to different harms based on their type of exposure. On the other hand, AI harms are also a consequence of a responsible entity's \textit{intent}. Responsible entities are classified as creators of AI if they invest resources into training an AI system. If a responsible entity uses or deploys an existing AI system with minimal training (e.g. an open source model or a commercially available product), we classify them as a deployer of AI. Creators of AI may intentionally \textit{create AI to harm}, but harm can also arise \textit{due to the creation} process of AI. Similarly, deployers of AI may intentionally \textit{deploy AI to harm}. But even in the absence of explicit intent, harm can arise \textit{from deployment}. In Figure~\ref{fig:ai_harm_framework} we visualise these relations as conceptual framework for modelling pathways to AI harms.

\section{Taxonomy of Harms}
\label{s:taxonomy}

We now present our taxonomy of harms, demonstrate how the taxonomy can be used to model pathways to harms, and define specific harms identified in speech generation incidents.


\subsection{Categorising Speech Generation Harms}

Following the approach laid out in Figure~\ref{fig:method}, we derived our taxonomy of harms by iteratively applying and revising the conceptual framework in Figure~\ref{fig:ai_harm_framework} to annotate the speech generation incidents in Table~\ref{tab:speech_gen_incidents} in Appendix~\ref{a:incidents}. Figure~\ref{fig:sg_harms_framework} in Appendix~\ref{a:taxonomy} shows how we expanded the conceptual framework to capture harm types and motives that we identified from speech generation incidents: affected entities that are the subject of speech generators can be \textit{compelled to speak} words they never said, or have their \textit{own voice taken}. An affected entity that interacts with a speech generator can be harmed because they are \textit{impacted} or \textit{deceived} by a synthetic voice. Affected entities that suffer due to speech generators may not interact with a speech generator at all, but still experience \textit{economic}, \textit{cultural}, \textit{physical} and \textit{social harms}, or \textit{harm to reputation or mind}. Affected entities that are excluded from speech generators may experience harms related to \textit{stereotyping}, \textit{erasure} and \textit{homogenisation}. 

Responsible entities that deploy speech generators with the intent of harm may do so to engage in \textit{fraud}, \textit{lawlessness}, \textit{trolling}, \textit{electoral sway} or \textit{polarisation}. Harms may also arise as unintended consequences of responsible entities acting negligently or pursuing a primary goal. Thus, we ascribe harms due to creation to underlying \textit{entertainment} or \textit{profiteering} motives (i.e. seeking to make excessive profits, irrespective of potential negative consequences for society or the environment), while harms from deployment can be traced back to \textit{featurisation} (i.e. adding new features to an existing product), \textit{narration} (i.e. adding a voice to narrate a viewpoint) and \textit{entertainment}. We found no incidents where speech generators were created to harm, and thus identified no further harm motives in this category. 

\subsection{Pathways to Harms}

\begin{figure*}[hbt]
    \centering
    \includegraphics[width=0.95\linewidth]{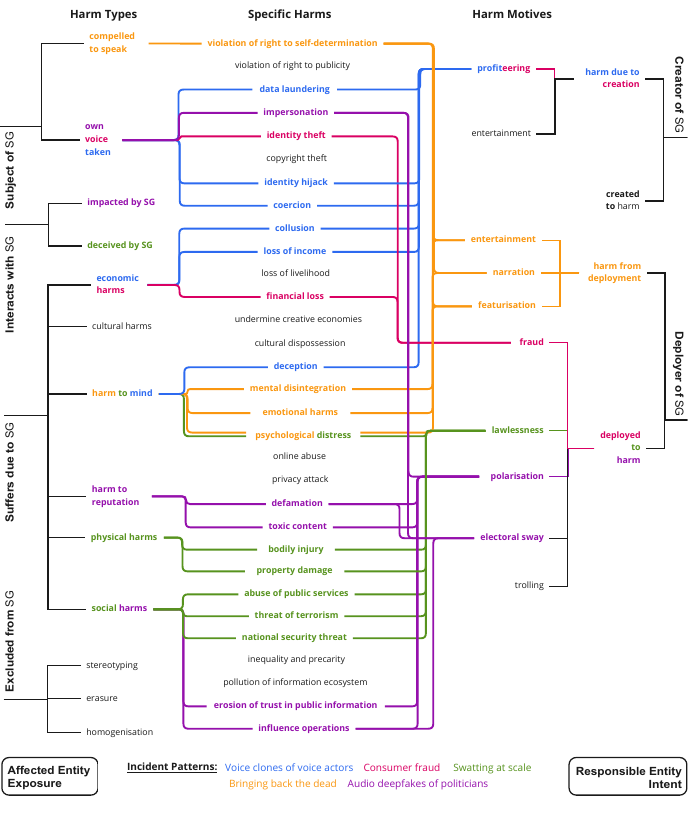}
    \caption{Pathways to harms are visualised by connecting specific harms to the exposure of an affected entity and the intent of a responsible entity. Colours represent incidents with common harm pathways based on harm types, motives and specific harms.}
    \label{fig:expanded_sg_harms_framework}
\end{figure*}

Using our taxonomy, we directly derived pathways to speech generator harms by observing patterns across incidents with similar affected entities, responsible entities, harm types an motives in Table~\ref{tab:speech_gen_incidents} in Appendix~\ref{a:incidents}. By connecting the harm types and exposure of affected entities with the intent and harm motives of responsible entities, we thus established pathways that lead to specific harms. Figure~\ref{fig:expanded_sg_harms_framework} shows five incident patterns that share common pathways to harm based on co-occurring stakeholders, harm types and motives. They are described next.

\subsubsection{\textbf{Voice clones of voice actors} (incidents 30 - 34) [\textcolor{blue}{blue}]}
Voice actors become the subject of speech generators when their own voices are taken by corporations to train and release AI models. Responsible entities are corporations with profiteering motives. We consider these incidents as identity hijacks, as voices are pursued for specific attributes (e.g. “a female voice with a North American accent”: incident 34), not the identity of the voice actor (e.g. Beverly Standing: incident 34). Responsible entities can collaborate with each other while blindsiding voice actors (incidents 30, 34). Tactics deployed by corporations are collusion, coercion, deception and data laundering. Even though the voice actors oftentimes signed contracts that permit unlimited AI use of their voice, these contracts were signed long before current generative AI (e.g. 2005, 2016 in incidents 31, 32), or without the voice actors' understanding of the implications. Voice actors describe the contracts as deceptive ("I did not realise") and coercive ("I had no choice"), and suffer economic harms from a loss of income. This pathway presents new specific harms created by speech generators.

\definecolor{cadmiumorange}{rgb}{0.93, 0.53, 0.18}
\subsubsection{\textbf{Bringing back the dead} (incidents 1 - 4) [\textcolor{cadmiumorange}{orange}]}
In several incidents deceased people have become the subject of speech generators by being compelled to speak with synthetic voices. We consider these incidents as a violation of the deceased's right to self-determination. Additionally, some incidents impersonate famous deceased people (incidents 3, 4). Responsible entities can be anonymous (incidents 1, 4) or corporations (US e-commerce giant: incident 2, South Korean broadcaster: incident 3). Their motives vary from entertainment to narration and featurisation. Beyond the deceased, affected entities can also be relatives  (incident 4) and audience members (incident 3) who interact with and are impacted by speech generators. While audience members may suffer no further harms, relatives can suffer emotional harms, mental disintegration and psychological distress. Speech generators create and exacerbate the specific harms in this pathway.

\definecolor{brightpink}{rgb}{1.0, 0.0, 0.5}
\subsubsection{\textbf{Consumer fraud} (incidents 15, 16) [\textcolor{brightpink}{pink}]}
Banks and tax offices increasingly mandate the use of voice authentication systems. In the incidents that we identified, journalists cloned their own voices to highlight how speech generator enabled identity theft increases vulnerabilities in voice authentication systems and can result in economic harms to consumers. Even though no actual harms occurred in these incidents, they demonstrate the risks that speech generators present to customers. In this harm pathway, responsible entities deploy speech generators to harm, with the motive of committing fraud. The incidents are facilitated by corporations with profiteering motives who sell voice cloning services without adequate safeguards to prevent misuse. This allows anybody to clone voices cheaply, easily, anonymously, in a short time and without technical know-how. Affected entities become the subject of speech generators and have their own voice taken. When fraudsters succeed, affected entities suffer financial loss due to speech generators. Speech generators exacerbate the specific harms in this pathway by providing tooling that scales the frequency of harms.

\definecolor{patriarch}{rgb}{0.5, 0.0, 0.5}
\subsubsection{\textbf{Audio deepfakes of politicians} (incidents 11, 13, 14) [\textcolor{patriarch}{purple}]}
Speech generators are increasingly used to spread manipulated information with audio deepfakes that impersonate political figures. Affected politicians become the subject of speech generators and have their own voice taken. When deployed with the intent to harm, audio fakes serve motives of polarisaton (incidents 11, 14) and electoral sway (incident 13). Unsurprisingly, responsible entities are anonymous. Messaging and social media platforms, which are unable to prevent the spread of deepfakes, facilitate this pathway. In some instances audio fakes spread toxic (incident 11) and defamatory (incidents 11, 13) content, thus causing the affected politician to also suffer reputational harms. Furthermore, citizens and society at large suffer social harms as audio fakes can be used to support influence operations and lead to the erosion of trust in public information. This pathway presents new specific harms created by speech generators.

\definecolor{forestgreen}{rgb}{0.13, 0.55, 0.13}
\subsubsection{\textbf{Swatting at scale} (incident 9) [\textcolor{forestgreen}{green}]}
Speech generators are used to incite police officers to conduct swatting attacks. On the one hand, police offers who interact with speech generators are affected entities, deceived by speech generators to believe bogus bomb threats or claims of civil disobedience. On the other hand, police officers who act on the calls are also responsible entities who perpetuate harm. The victims of the swatting attacks, who do not interact with speech generators, may suffer injury, property damage, and psychological distress. When conducted at scale, swatting attacks can be used to overload, misdirect or misuse policing services. They thus present a threat of terrorism, a national security threat and an abuse of public services. Citizens carry the burden of these social harms when they materialise. The responsible entities who instigate swatting attacks are anonymous and intentionally deploy speech generators to harm in the pursuit of lawlessness. Speech generators offer responsible entities anonymity and increase the frequency with which swatting can be instigated. They thus exacerbate the specific harms in this pathway.

\subsection{Specific Harms}

With the high-level categorisation of harm types and motives established, and several pathways to harm demonstrated, we next discuss the specific harms from incidents listed in Table~\ref{tab:specific_harms_descriptions}.

\subsubsection{\textbf{Subject of} speech generator harms}
The \textit{violation of right to self-determination} harm (incidents 1 - 4) arises when an individual's autonomy and decision-making rights about their identity or personhood are diminished, irrespective of whether they are alive or dead. By contrast, the \textit{violation of right to publicity} harm (incidents 22 - 25, 35) occurs when an entity's identifiable aspects, like their name or voice, are exploited for commercial gain. 

\renewcommand{\arraystretch}{1.6}
\begin{table*}[!hbt]
\footnotesize
\caption{Taxonomy for categorising speech generation harms.}
\label{tab:specific_harms_descriptions}
\vspace{-1.2em}
\begin{tabular}{|L{0.12\textwidth}L{0.25\textwidth}|L{0.6\textwidth}|} \hline
\textbf{Harm type} & \textbf{Specific harm} & \textbf{Description} \\ \hline
\multicolumn{3}{c}{\textbf{Affected entity \textit{subject of} speech generator}} \\ \hline
Compelled to speak & Violation of right to self-determination & Revoking an individual's control and autonomy in decision making processes that concern their identity or personhood while they are alive or after their death. \\ 
 & Violation of right to publicity & Using an entity’s name, likeness, or other recognisable aspects of their persona, such as their voice, for commercial purposes. \\ \hline
& Data laundering & Non-consensual data collection~\cite{Shelby2022SociotechnicalReduction}, exploitative data sourcing and enrichment~\cite{Weidinger2023SociotechnicalSystems}. \\
 & Impersonation & Pretending to be a specific person for entertainment or fraud. \\
Own voice taken & Identity theft & Copy or clone someone's likeness in order to gain from it through criminal activity. \\
 & Copyright theft & Lay claim to somebody's creative work, without attribution, consent or compensation. \\
 & Identity hijack & Appropriate someone's likeness, without care for their identity, in order to gain from it. \\ 
 & Coercion & Affected entities face the choice of suffering loss, or of unconditionally accepting the use of their personal data or creative work, oftentimes without compensation. \\ \hline
 \multicolumn{3}{c}{\textbf{Affected entity \textit{suffers due to} speech generator}} \\ \hline 
& Collusion & Affected entities fall victim to collusive practices of industry stakeholders. \\
Economic harms & Loss of income & Reduction in the ability to gain financially from one's work. \\
 & Loss of livelihood & Arises when an individual's ability to generate income from their activities or profession is impacted, and may include an associated loss of identity, community and income. \\
 & Financial loss & Loss of money owned by an affected entity due to the wrong-doing of others. \\ \hline
Cultural harms & Undermine creative economies & Devaluing human creativity, artistic expression and imagination (derived from~\cite{Weidinger2023SociotechnicalSystems}). \\
 & Cultural dispossession & Intentional and unintentional erasure of any cultural goods and values, like ways of speaking, expressing humour, or sounds and voices that contribute to a national identity. \\ \hline
Physical harms & Bodily injury & Encouraging harm to body or actions that lead to injury. \\ 
 & Property damage & Provoking actions that lead to damage of material goods. \\ \hline
 & Abuse of public services & Overloading, misdirecting or misusing services that governments provide to citizens to address their needs. \\
 & Threat of terrorism & Risk of unlawful use of violence and intimidation against civil society in order to accomplish political aims. \\
Social harms & National security threat & Actions of malevolent actors that threaten the sovereignity, territorial integrity, constitutional order and social stability of a sovereign state. \\
 & Inequality and precarity & Amplifying social and economic inequality, or precarious or low-quality work~\cite{Weidinger2023SociotechnicalSystems}. \\
 & Pollution of information ecosystem & Contaminating publicly available information with false or inaccurate information~\cite{Weidinger2023SociotechnicalSystems} \\
 & Erosion of trust in public information & Diminishing trust in public information and knowledge~\cite{Weidinger2023SociotechnicalSystems} \\
 & Influence operations & Facilitating large-scale disinformation and targeted manipulation of public opinion~\cite{Weidinger2023SociotechnicalSystems} \\ \hline
Harm to reputation & Defamation & Facilitating slander, defamation or false accusations~\cite{Weidinger2023SociotechnicalSystems} that damage the reputation of an affected entity. \\
 & Toxic content & Generating content that violates community standards, including harming or inciting hatred or violence against individuals and groups~\cite{Weidinger2023SociotechnicalSystems} \\ \hline
 & Deception & Affected entities are tricked into a belief or to commit an action. \\
 & Mental disintegration & Affects a person's ability to process their emotions and to retain a healthy mental state, leading to a diminished ability to make sense of reality. \\ 
Harm to mind & Emotional harms & Injury to a person's feelings, or stimulating negative feelings. \\
 & Psychological distress & Induces a state of emotional suffering of significant severity, which can include anxiety, panic, trauma, depression and paranoia. \\
 & Online abuse & Various forms of online harassment and cyberbullying. \\
 & Privacy attack & Personal, private or sensitive information is made publicly available. \\ \hline
 \multicolumn{3}{c}{\textbf{Affected entity \textit{interacts with} speech generator}} \\ \hline
Impacted by & \textit{- no specific harms identified -} & Affected entity is impacted by their interaction with a speech generator. \\ \hline
Deceived by & \textit{- no specific harms identified -} & Affected entity is mislead by their interaction with a speech generator. \\ \hline
\multicolumn{3}{c}{\textbf{Affected entity \textit{excluded from} speech generator}} \\ \hline
Stereotyping & \textit{- no incidents -} & Representation in a stylised manner based on certain personal attributes. \\ \hline
Erasure & \textit{- no incidents -} & Not represented. \\ \hline
Homogenisation & \textit{- no incidents -} & Representation without distinguishing individual uniqueness. \\ \hline
\end{tabular}
\end{table*}

\textit{Data laundering} (incidents 30 - 34) occurs when responsible entities engage in non-consensual or exploitative data collection. This can involve \textit{copyright theft} (incidents 21 - 25, 35), which relates to the exploitation of creative work without proper acknowledgment or consent. \textit{Impersonation} (incidents 3 - 8, 10 - 14, 19, 20, 22 - 28) is the act of assuming someone else's persona for entertainment or fraud. This is different from but can co-occur with \textit{identity theft} (incidents 15, 16, 19, 20, 26 - 28), in which someone's likeness is explicitly copied to conduct criminal activities. 
We distinguish \textit{identity hijack} (incidents 29 - 35) from identity theft, to encompass the class of incidents where someone's likeness is appropriated for commercial gain, without regard for their individual identity. This typically occurs when people's voices are used to train speech generators, or when voice actors find synthetic versions of their voice sold or distributed on the internet. While these synthetic voice clones are not explicitly associated with their identity, affected entities did not consent to this use, and rarely received adequate compensation for it. Finally, when responsible entities use \textit{coercion} (incidents 30 - 34) as a tactic, they pressure affected entities to accept unfavourable conditions for the use of their personal data or creative work.

\subsubsection{\textbf{Suffers due to} speech generator harms}
The majority of specific harms occur when an affected entity suffers due to speech generators. Harms of \textit{collusion} (incidents 30 - 34) occur when responsible entities cooperate to gain an unfair advantage to the detriment of individuals. Affected entities who suffer \textit{loss of income} (incidents 22 - 24, 30 - 34) have a reduced ability to gain financially from their work. \textit{Loss of livelihood} (incident 29) occurs when an affected entity can no longer practice their profession. This is more existential and permanent than loss of income, and may have consequences beyond financial loss, such as loss of identity or community that an affected entity gained from their profession. \textit{Financial loss} (incidents 15, 16, 19, 20, 26, 27) concerns the loss of money, but is not specific to income or work. The \textit{undermining of creative economies} (incidents 21, 22, 25,29, 30) is associated with a devaluation of human creativity, artistic expression and imagination. It affects entire industries, such as voice acting, or music production. This is a broader definition than ``substituting original [creative] works with synthetic ones" which was previously proposed~\cite[p. 31]{Weidinger2023SociotechnicalSystems}. \textit{Cultural dispossession} (incident 29) involves the erasure of cultural goods and values of any nature. \textit{Bodily injury} and \textit{property damage} (incident 9) are physical harms that arise when speech generators are used to encourage and provoke harm to a person's body and material goods.

Specific social harms share the commonality that they affect citizens and society at a large scale. Frequently, these harms are not immediately recognisable after a single incident, and only become visible over time, after many incidents, when incidents have regional spread or occur at very particular moments in time. \textit{Abuse of public services}, \textit{threat of terrorism} and \textit{national security threats} (incident 9) arise when government services are intentionally overloaded, misdirected or misused, when violence is used against citizens for political aims and when malicious actors threaten state sovereignty, integrity and order. \textit{Inequality and precarity} (incident 29) may arise when economic harms such as \textit{loss of income} and \textit{loss of livelihood} prevail for extended periods of time and affect a community of people. \textit{Pollution of information ecosystems} (incident 17), \textit{erosion of trust in public information} (incidents 10 - 12, 14, 17, 18) and \textit{influence operations} (incidents 11, 13, 14, 17) relate to the fabrication and spread of information for malicious purposes, and subsequent social mistrust of public information. In incidents where we identified \textit{defamation} (incidents 10, 11, 13) and \textit{toxic content} (incidents 6, 8, 10, 11, 13), we categorised them as harms to reputation, as speech generators have been used to make well known people say offensive or inflammatory things.  

\textit{Deception} (incidents 7, 18, 30 - 34) occurs when responsible entities trick or mislead affected entities into believing something, or into taking an action.  We consider \textit{mental disintegration} (incidents 1, 2) as diminishing the capacity of an affected entity to make sense of reality, for example by reducing an individual's ability to distinguish whether they are talking with a living or deceased relative. \textit{Emotional harms} (incidents 1, 2, 4), by contrast, result in injury to a person's feelings. The consequences of \textit{psychological distress} (incidents 1, 26 - 28) are more severe and persistent than emotional harms. By nature of being non-physical, we categorise \textit{online abuse} (incidents 5, 6) and \textit{privacy attacks} (incident 5) as harm to mind.

\subsubsection{\textbf{Interacts with} and \textbf{excluded from} speech generator harms}
There are no specific harms associated with affected entities interacting with or being excluded from speech generators. For \textit{interacts with} harms, the type of harm was the most fine-grained categorisation that we identified from incidents. Affected entities can be impacted by (incidents 1 - 4, 6, 10 - 14, 17) or deceived by (incidents 7, 9, 18 - 20, 26 - 28) speech generators. Harms from interaction often lead to affected entities also suffering due to speech generators, or to inflicting harm on others. For example, deceived grandparents who believe that their grandchild has just been in an accident subsequently suffer financial loss and psychological distress (incidents 26, 27). However, interaction alone does not imply that an affected entity also suffers due to speech generators. We identified no incidents with harms due to affected entities being excluded from speech generators, and thus did not categorise specific harms as stereotyping, erasure and homogenisation. It is likely that specific harms in these categories will be identified in future as the use of speech generation technology in products increases.





\section{Discussion}
\label{s:discussion}


We set out to develop a harm taxonomy that is based on an explicit categorisation framework, that supports modelling pathways to AI harms and that captures specific harms of speech generators. In Section~\ref{s:framework}, we presented a conceptual framework for categorising generative AI harms based on the exposure of an affected entity, and the intent of a responsible entity. Then, in Section~\ref{s:taxonomy}, we adapted this framework to the speech generation domain to develop a taxonomy of harms of speech generators. We illustrated how the taxonomy can be used to model pathways to harm from patterns in related speech generation incidents. We now compare our proposed taxonomy against existing taxonomies, propose directions for future work, and discuss limitations of our approach and the resultant taxonomy.

\subsection{Comparison against Existing Taxonomies}

In the development of our taxonomy we drew inspiration from prior research, in particular the taxonomies of \citet{Shelby2022SociotechnicalReduction} and \citet{Weidinger2023SociotechnicalSystems}. We adopted the specific harms and harm types identified in these taxonomies as far as possible. Table~\ref{tab:taxonomy_extension} in Appendix~\ref{a:taxonomy} offers a comparison between our taxonomy and these two taxonomies. Shelby et al.'s~\cite{Shelby2022SociotechnicalReduction} taxonomy captures an extensive list of specific harms where affected entities suffer due to AI. A number of these harms were not mentioned in speech generation incidents, and are thus not represented in our taxonomy. In return, this taxonomy lacks harm types and specific harms for affected entities who are the subject of, or interact with AI. This is unsurprising, as the taxonomy does not focus on generative AI, where these types of exposure risks are more prominent. By contrast, Weidinger et al.'s~\cite{Weidinger2023SociotechnicalSystems} taxonomy emphasises harms that stem from the interaction of end users with large language models (LLMs). With LLMs, interaction is a key enabler of harms, which may explain the dominance of these harm types. Again, this set of specific harms was not covered in our taxonomy.  

This comparison highlights how our bottom-up, incident based approach compliments existing harm taxonomies. Table~\ref{tab:taxonomy_extension} also illustrates how incorporating specific harms from previous taxonomies can extend this work beyond speech generators to predictive AI systems and LLMs. For example, Shelby et. al's specific harms expand our economic, cultural, physical and social harm types, and have led us to add two new harm types, \textit{participation harms} and \textit{misrepresentation harms}. Similarly, we categorised Weidinger et. al's harms into five new harm types that can arise when an affected entity interacts with AI: affected entities can be \textit{manipulated} or \textit{persuaded by}, \textit{receive dangerous} or \textit{false information from}, or become \textit{reliant on} AI. 

We hope that this comparison is useful for others who wish to adapt our taxonomy and conceptual framework to study AI harms in new domains. Particularly, we believe that our framing presents a promising point of departure for categorising harms in other multimodal generative AI systems, such as image and video generation. For example, specific harms arising in incidents that involve the non-consensual generation of pornographic imagery~\cite{moreau2024fake} can be categorised as affected entities being the \textit{subject of} \textit{devious depictions} or having their \textit{face and body image taken}, as opposed to being \textit{compelled to speak} and having their \textit{own voice taken}. Similarly, factually incorrect depictions of history~\cite{robertson2024google} can be categorised as \textit{misrepresentation} and \textit{erasure} harms, through which affected entities are \textit{excluded from} AI.


\subsection{Future Research}


This research highlights the diverse risks associated with speech generators, and the need to understand how harms arise from their creation and deployment. The ultimate objective for doing this is to effectively mitigate these risks and prevent adverse consequences. To address the risks posed by speech generators, it is necessary to distinguishing between testable harms (e.g. receiving dangerous or false information from a model) and non-testable harms (e.g. abuse of public services, financial loss and pollution of information ecosystem). Testable harms can be measured by conducting rigorous capability evaluations that characterise model outputs. While this is increasingly done for LLMs~\cite{vidgen2024introducing}, it is not yet common for speech generators. Instead, current evaluations rely on listener tests, which are subjective and oftentimes unreliable~\cite{wagner_speech_2019, kirkland_stuck_2023}. For speech generators, further research is required to determine which capabilities can and should be measured to assess harmful behaviour, how these capabilities can be operationalised and measured, and to identify potential shortcomings of capability evaluations. 

Non-testable harms often relate to system-level risks that have an extensive scope, that occur over a long time horizon, that affect society at large and that are not directly identifiable from model outputs. Test-based capability evaluations are insufficient for identifying and mitigating risks that lead to these kinds of harms. Here, harm pathways present a promising direction for analysing the emergence and propagation of harm. Future research should thus study how harm pathways emerge, and their dependence on affected and responsible entities, as well as mediating factors. For example, while testing interaction related harm types is necessary for categorising, evaluating and moderating content created by LLMs, further research is required to understand how an affected entity who interacts with a LLM responds to the content, what makes them vulnerable to it, and how specific harms, like polluted information ecosystems or the erosion of trust in public information, arise from this. Across testable and non-testable harms, future research should also study bias in model outputs to understand the extent to which demographic and other personal attributes of affected entities impact their harm exposure.


Finally, it is necessary to develop and validate approaches to mitigate testable and non-testable harms. Some recent advances have shown promising directions. For example, AntiFake provides a proactive, technical approach to prevent impersonation and identity theft through unauthorised speech synthesis~\cite{zhiyuan2023antifake}. Affected entities can use AntiFake to perturb speech data that they make publicly available, thus preventing themselves from having their \textit{own voice taken} and becoming the \textit{subject of} speech generators. An alternative, defensive technical mitigation strategy is presented by works such as WavMark~\cite{chen2024wavmark} and AudioSeal~\cite{roman2024proactive}, which propose watermarking for generated speech. These interventions are aimed at deterring responsible entities from propagating \textit{harm from deployment} or systems that are \textit{deployed to harm}, by making it possible to detect content that has been generated by a particular model.

Regulatory, policy and civil society interventions that seek to develop institutional strategies to mitigate generative AI harms can benefit from considering harm pathways to assign accountability, identify gaps, and analyse the efficacy of mitigation strategies. For example, harm from deployment can oftentimes be attributed to well-intended developers and product teams who work in legitimate organisations. Responsible design practices~\cite{pai2023} that include transparency and disclosure, product reviews and approval thus present a potential mitigation approach. On the other hand, when speech generators are deployed to harm, responsible entities are frequently anonymous. However, these harms are facilitated by technology vendors and an open-source ecosystem~\cite{widder2022limits} that make model access and adaptation easy, cheap and frictionless. Potential mitigation approaches that increase friction by raising the technical barrier to entry for voice cloning and voice conversion models, safety-oriented strategies for model release (e.g. one (default) voice only), and mechanisms for tighter access control (e.g. for multi-speaker models) are avenues to explore in future work. Where harm arises due to creation, this is often linked with ruthless profit drives of businesses. In these instances, future work should explore accountability mechanisms, and how affected entities can coordinate efforts to approach responsible entities collectively. Some promising efforts in this direction are work on certification for training data sourcing~\cite{fairlytrained2024} and opt-out mechanisms from training data for creators~\cite{hibt2024}.

\subsection{Limitations}


There exist many pathways to harm, and consequently also a variety of ways of categorising the specific harms in our taxonomy. Other researchers may choose to categorise specific harms differently to us. While we motivate our choices with our analysis of incidents, this has two shortcomings. Firstly, the incidents we consider are limited by the databases we searched and by our filtering strategy. Secondly, even if we were to be able to discover all relevant incidents that ever occurred, then a harm still needs to have happened before we can capture its pathway. It is desirable to study modelling approaches that can preempt pathways to harm, so that mitigating actions can be taken even before harms materialise. 

It is unlikely that any taxonomy of harms will ever be complete, including our own. New harms emerge over time as system capabilities and vulnerabilities become better known and exploited. As our taxonomy is derived from reported incidents, it does not cover incidents of harm that have not been reported. For example, there are communities that may benefit significantly from speech generation technologies (e.g. people with speech impediments). If speech generation systems have not been designed with their needs in mind, and consequently do not work for them, they can be subjected to stereotyping, homogenisation and erasure harms. While our taxonomy alludes to the presence of these harms in the \textit{excluded from} category for affected entity exposure, we did not cover specific harms of these types, which remains an area for future work. 

Finally, we do not assign intent to current AI systems, and thus did not consider them as responsible entities. Rather, our framework looks beyond the technology, to human and institutional stakeholders who are responsible for harms perpetuated by AI systems.

\section{Conclusion}
\label{s:conclusion}

Speech generators threaten income sources and the creative works of voice actors and musicians, have been used to steal millions from individuals and companies, are stoking political unrest, and are fuelling physical violence through new waves of swatting attacks. To mitigate risks of speech generators, this paper positions that it is necessary to understand \textit{what risks exist \underline{and} how risks arise} from stakeholder interactions with generative AI. By analysing speech generation incidents, we found that patterns of specific harms depend on whether individuals are a subject of, interact with, suffer due to, or are excluded from speech generators. At the same time, we found that specific harms are also a consequence of the motives of the creators and deployers of the systems. Based on these insights we proposed a conceptual framework for modelling pathways to ethical and safety harms of generative AI, which we used to develop a taxonomy of harms of speech generators. Our relational approach captures the complexity of risks and harms in sociotechnical AI systems, and yields a taxonomy that can support appropriate policy interventions and decision making for the responsible development and release of speech generation models.

\begin{acks}
We thank Tiffany Georgievski for contribution to and discussion of AI incidents, William Thong for helpful feedback and discussions on the conceptual framing and specific harms, Josh Meyer for discussions on ethical considerations for speech generation systems and the volunteers who have built and continue to contribute to the various AI incident databases.
\end{acks}

\bibliographystyle{ACM-Reference-Format}
\bibliography{references, manual_references}

\clearpage
\onecolumn
\appendix
\section{Appendix}
\label{appendix}




\subsection{Annotated Speech Generation Incidents}
\label{a:incidents}

\begin{footnotesize}
\begin{longtable}{C{0.02\textwidth}L{0.25\textwidth}L{0.1\textwidth}L{0.1\textwidth}L{0.1\textwidth}L{0.11\textwidth}L{0.085\textwidth}L{0.14\textwidth}}
\caption{Incidents of harm from synthetically generated speech. AE: Affected entity, RE: Responsible entity}\\
\label{tab:speech_gen_incidents}
\textbf{id} & \textbf{Summary of incident} & \textbf{AE: \newline{subject of}} & \textbf{AE: \newline{interacts with}} & \textbf{AE: \newline{suffers due to}} & \textbf{RE intent} & \textbf{RE type} & \textbf{Specific harms} \endhead \hline
1 & Fake videos of dead children that narrate their own murders and the abuse and violence they suffered at the hands of their perpetrators have been spread on social media.\footnote{\url{https://www.express.co.uk/news/uk/1771516/tiktok-ai-videos-murdered-children}} & deceased: given a voice & relatives: impacted & relatives: harm to mind & harm from deployment: narration & anonymous & violation of right to self-determination; psychological distress; mental disintegration; emotional harms; \\
2 & Amazon plans to let people turn their dead loved ones’ voices into digital assistants.\footnote{\url{https://www.theguardian.com/technology/2022/jun/23/amazon-alexa-could-turn-dead-loved-ones-digital-assistant}} & deceased: own voice taken; given a voice & relatives: impacted & relatives: harm to mind & harm from deployment: featurisation & corporation & violation of right to self-determination; mental disintegration; emotional harms \\
3 & National broadcaster in South Korea creates a show with new songs sung with the cloned voice of deceased musician Kim Kwang-seok, who took his own life at age 31.\footnote{\url{https://edition.cnn.com/2021/01/25/asia/south-korea-kim-kwang-seok-ai-dst-hnk-intl/index.html}} & deceased: own voice taken; given a voice & audience: impacted & - & harm from deployment: entertainment & corporation & violation of right to self-determination; impersonation \\
4 & Voice of deceased actor Robin Williams has been recreated to the dismay of his daughter Zelda.\footnote{\url{https://www.independent.co.uk/arts-entertainment/films/news/robin-williams-ai-voice-daughter-b2422506.html}} & deceased: own voice taken; given a voice & relatives: impacted & relatives: harm to mind & harm from deployment: entertainment & anonymous & violation of right to self-determination; impersonation; emotional harms \\
5 & Online trolls used ElevenLabs to clone voice actors' voices to harass them. The voices read out victims' home addresses and post the results online.\footnote{\url{https://www.vice.com/en/article/93axnd/voice-actors-doxed-with-ai-voices-on-twitter}} & public figure: own voice taken & - & public figure: harm to mind & deployed to harm: trolling & anonymous & impersonation; online abuse; privacy attack; \\
6 & 4chan members cloned voices of celebrities to disseminate violent, transphobic, homophobic and racist content.\footnote{\url{https://www.vice.com/en/article/dy7mww/ai-voice-firm-4chan-celebrity-voices-emma-watson-joe-rogan-elevenlabs}} & public figure: own voice taken & audience: impacted & public figure: harm to reputation; audience: harm to mind  & deployed to harm: trolling & anonymous & impersonation; online abuse; toxic content \\
7 & TikTok user cloned imprisoned influencer Andrew Tate to convince followers that Tate is free, posting videos; and giving self-help advice.\footnote{\url{https://www.vice.com/en/article/5d3n8z/a-scammer-is-pretending-to-be-andrew-tate-on-tiktok-and-racking-up-millions-of-views}} & public figure: own voice taken & audience: deceived & - & harm due to creation: profiteering & individual & impersonation; deception \\
8 & A website enabled anybody to make a clone of public figure and Canadian Professor Jordan Peterson say anything.\footnote{\url{https://www.vice.com/en/article/43kwgb/not-jordan-peterson-voice-generator-shut-down-deepfakes}} & public figure: own voice taken & - & public figure: harm to reputation & harm from deployment: entertainment & anonymous & impersonation; toxic content \\
9 & Nationwide swatting wave in the US can be traced to a swatting-as-a-service account on Telegram that uses synthesized voices to pressure law enforcement to specific locations.\footnote{\url{https://www.vice.com/en/article/k7z8be/torswats-computer-generated-ai-voice-swatting}} & - & police: deceived & victim: physical harms, harm to mind; citizens: social harms  & deployed to harm: lawlessness & anonymous & abuse of public services; national security threat; threat of terrorism; bodily injury; property damage; pyschological distress \\
10 & Fake audio captures the President of Japan making vulgar statements.\footnote{\url{https://japannews.yomiuri.co.jp/politics/politics-government/20231104-147695/}} & politician: own voice taken & citizens: impacted & politician: harm to reputation; citizens: social harms  & harm from deployment: entertainment & individual & impersonation; erosion of trust in public information; toxic content; defamation \\
11 & Fake audio captures the President of the US making derogatory comments about transgender people.\footnote{\url{https://www.reuters.com/article/factcheck-biden-transphobic-remarks/fact-check-video-does-not-show-joe-biden-making-transphobic-remarks-idUSL1N34Q1IW}} & politician: own voice taken & citizens: impacted & politician: harm to reputation; citizens: social harms  & deployed to harm: social polarisation & anonymous & impersonation; erosion of trust in public information; influence operations; toxic content; defamation \\
12 & Fake audio imitates Omar al-Bashir, former leader of Sudan, adding confusion and instability in war-torn Sudan.\footnote{\url{https://www.bbc.co.uk/news/world-africa-66987869}} & politician: own voice taken & citizens: impacted & citizens: social harms & harm from deployment: narration & individual & impersonation; erosion of trust in public information; \\
13 & Fake audio of Slovak political leader \& journalist discussing how to rig the elections adds political instability two days before elections.\footnote{\url{https://www.wired.com/story/slovakias-election-deepfakes-show-ai-is-a-danger-to-democracy/}} & politician: own voice taken & citizens: impacted & citizens: social harms & deployed to harm: electoral sway & anonymous & impersonation; influence operations, defamation \\
14 & Fake audio of Mayor of London calls for Armistice Day to be rescheduled to accommodate a pro-Palestinian march.\footnote{\url{https://www.bbc.co.uk/news/uk-england-london-67389609}} & politician: own voice taken & citizens: impacted & citizens: social harms & deployed to harm: social polarisation & anonymous & impersonation; erosion of trust in public information; influence operations \\
15 & Australian journalist broke into their own government self-service account to highlight the vulnerability of the Australian Bank \& Tax Office to voice attacks.\footnote{\url{https://www.theguardian.com/technology/2023/mar/16/voice-system-used-to-verify-identity-by-centrelink-can-be-fooled-by-ai}} & customer: own voice taken & - & customer: economic harms & deployed to harm: fraud & individual & identity theft; financial loss \\
16 & Journalist used voice cloning to trick voice ID systems of banks in the EU and US.\footnote{\url{https://www.vice.com/en/article/dy7axa/how-i-broke-into-a-bank-account-with-an-ai-generated-voice}} & customer: own voice taken & - & customer: economic harms & deployed to harm: fraud & individual & identity theft; financial loss \\
17 & Human avatars with generated voices spread propaganda to divert attention and change political narratives in Venezuela.\footnote{\url{https://english.elpais.com/international/2023-02-22/theyre-not-tv-anchors-theyre-avatars-how-venezuela-is-using-ai-generated-propaganda.html}} & - & citizens: impacted & citizens: social harms & deployed to harm: electoral sway & politician & erosion of trust in public information; influence operations; pollution of information ecosystem \\
18 & Some citizens feel mislead by New York's mayor, who intentionally used audio deepfakes to promote local events to residents by robocalling them in their home languages, which he doesn't speak.\footnote{\url{https://news.sky.com/story/new-yorks-mayor-uses-audio-deepfakes-to-call-residents-in-languages-he-doesnt-speak-12986816}} & - & citizens: deceived & citizens: social harms & deployed to harm: electoral sway & politician & deception; erosion of trust in public information \\
19 & CEO of UK based energy company scammed by fake audio of his boss's voice and paid EUR 200k to fraudsters.\footnote{\url{https://www.wsj.com/articles/fraudsters-use-ai-to-mimic-ceos-voice-in-unusual-cybercrime-case-11567157402}} & employee: own voice taken & employee: deceived & organisation: economic harms & deployed to harm: fraud & anonymous & impersonation; identity theft; financial loss \\
20 & Branch manager of a Japanese company based in Hong Kong scammed by fake audio of his boss's voice and paid \$35mil to fraudsters.\footnote{\url{https://www.forbes.com/sites/thomasbrewster/2021/10/14/huge-bank-fraud-uses-deep-fake-voice-tech-to-steal-millions/}} & employee: own voice taken & employee: deceived & organisation: economic harms & deployed to harm: fraud & anonymous & impersonation; identity theft; financial loss \\
21 & Anti-piracy group RIAA subpoenas Discord to shut down AI Hub server due to allegations of copyright infringement.\footnote{\url{https://torrentfreak.com/riaa-targets-ai-hub-discord-users-over-copyright-infringement-230622/}} & artist: own voice taken & - & industry: cultural, economic harms & harm due to creation: entertainment / profiteering & individual & copyright theft; undermine creative economies \\
22 & Voice cloning and audio generation were used to create "Heart on my sleeve";  a mixup with the voices of Drake and the Weeknd.\footnote{\url{https://variety.com/2023/music/news/fake-ai-generated-drake-weeknd-collaboration-heart-on-my-sleeve-1235585451/}} & artist: own voice taken & - & artist: economic harms; industry: cultural harms  & harm due to creation: entertainment / profiteering & anonymous & copyright theft; impersonation; undermine creative economies; violation of right to publicity; loss of income \\
23 & YouTube channel features content using the voice of celebrity Jay-Z to reproduce poems and songs without his consent and despite his demand for them to be removed.\footnote{\url{https://www.theverge.com/2020/4/28/21240488/jay-z-deepfakes-roc-nation-youtube-removed-ai-copyright-impersonation}} & artist: own voice taken & - & artist: economic harms & harm due to creation: entertainment / profiteering & individual & impersonation; violation of right to publicity; loss of income \\
24 & Singaporean singer replaced by an AI clone of herself without her consent, against her will and in detriment of her own music career.\footnote{\url{https://www.digitalmusicnews.com/2023/05/30/singaporean-singer-stefanie-sun-career-hijacked-ai/}} & artist: own voice taken & - & artist: economic harms & harm due to creation: entertainment / profiteering & anonymous & impersonation; violation of right to publicity; loss of income \\
25 & Voice cloning and audio generation was used to create a track with musician Bad Bunny's voice, which the artist explicitly rejected.\footnote{\url{https://people.com/bad-bunny-is-furious-about-an-ai-track-using-his-voice-8399608}} & artist: own voice taken & - & industry: cultural harms & harm due to creation: entertainment & anonymous & violation of right to publicity; impersonation; undermine creative economies \\
26 & Senior citizens in Newfoundland, Canada got scammed to pay money to fraudsters who used generated voices to pretend they are their grandchildren who had just been in an accident (8 separate incidents).\footnote{\url{https://www.cbc.ca/news/canada/newfoundland-labrador/ai-vocal-cloning-grandparent-scam-1.6777106}} & grandchild: own voice taken & relative: deceived & relative: economic harms, harm to mind & deployed to harm: fraud & anonymous & impersonation; identity theft; psychological distress; financial loss \\
27 & Elderly couple in Regina, US, nearly got scammed to pay fraudsters who claimed to be their grandson who had just been in a car accident and needed help.\footnote{\url{https://leaderpost.com/news/local-news/regina-couple-says-possible-ai-voice-scam-nearly-cost-them-9400}} & grandchild: own voice taken & relative: deceived & relative: economic harms, harm to mind & deployed to harm: fraud & anonymous & impersonation; identity theft; psychological distress; financial loss \\
28 & US mother threatened with kidnapping of her daughter in the hope of extorting ransom.\footnote{\url{https://edition.cnn.com/2023/04/29/us/ai-scam-calls-kidnapping-cec/index.html}} & child: own voice taken & relative: deceived & relative: economic harms, harm to mind & deployed to harm: fraud & anonymous & impersonation; identity theft; psychological distress \\
29 & Foreign-owned voice cloning companies are leaving Latin American voice actors without work or income.\footnote{\url{https://restofworld.org/2023/ai-voice-acting/}} & artists: own voice taken & - & artists: economic harms, Global South community: cultural harms, social harms & harm due to creation: profiteering & corporation & identity hijack; loss of livelihood; cultural dispossession; undermine creative economies; inequality and precarity \\
30 & Spotify’s audiobook narrators signed contracts that unknowing to them permitted Apple to use audiobook files for training AI models.\footnote{\url{https://goodereader.com/blog/audiobooks/audiobook-narrators-and-authors-fear-apple-using-their-voices-to-train-ai}} & artist: own voice taken & - & artist: economic harm & harm due to creation: profiteering & corporation & identity hijack; loss of income; coercion; collusion; data laundering; deception \\
31 & Voice over artist Greg Marston signed a contract in 2005 that unknowingly to him permits unlimited AI use of his voice. He now finds his voice sold as a service on the websites of third-party sites who have bought his voice clone from IBM.\footnote{\url{https://www.ft.com/content/07d75801-04fd-495c-9a68-310926221554}} & artist: own voice taken & - & artist: economic harm & harm due to creation: profiteering & corporation & identity hijack; loss of income; coercion; collusion; data laundering; deception \\
32 & Voice over artist Mike Cooper signed a contract in 2016 that unknowingly to him permits unlimited AI use of his voice. He now finds his voice sold as a service without his consent and without compensation.\footnote{\url{https://www.mikecoopervoiceover.com/behind-the-mike/2023/3/11/send-in-the-clones}} & artist: own voice taken & - & artist: economic harm & harm due to creation: profiteering & corporation & identity hijack; loss of income; coercion; collusion; data laundering; deception \\
33 & Voice actor Remie Clarke signed a contract in 2020 that unknowing to her permits unlimited AI use of her voice. She now finds her voice across the Internet without her consent and without compensation.\footnote{\url{https://www.independent.ie/videos/irish-voiceover-artist-losing-out-after-finding-ai-version-of-her-voice-being-sold-online/a683695072.html}} & artist: own voice taken & - & artist: economic harm & harm due to creation: profiteering & corporation & identity hijack; loss of income; coercion; collusion; data laundering; deception \\
34 & Unknown to her, voice actor Beverly Standing's voice has been sold on to TikTok  by a professional association who previously contracted her. TikTok users can now add her voice to generate synthetic speech with a North American accent for any content. \footnote{\url{https://voicebot.ai/2021/05/07/voice-actor-sues-tiktok-for-generating-text-to-speech-voice-from-her-recordings-without-permission/}} & artist: own voice taken & - & artist: economic harm & harm due to creation: profiteering & corporation & identity hijack; loss of income; coercion; collusion; data laundering; deception \\
35 & Actor Stephen Fry's voice has been cloned without his consent and used to generate the voice-over of a documentary film.\footnote{\url{https://www.standard.co.uk/news/tech/stephen-fry-voice-ai-cloned-documentary-b1108629.html}} & artist: own voice taken & - & industry: cultural harms & harm due to creation: profiteering & corporation & identity hijack; copyright theft;violattion of right to publicity; undermine creative economies
\end{longtable}
\end{footnotesize}

\subsection{Taxonomy of Harms of Speech Generators}
\label{a:taxonomy}

\begin{figure}[hbt]
    \centering
    \includegraphics[width=0.9\textwidth]{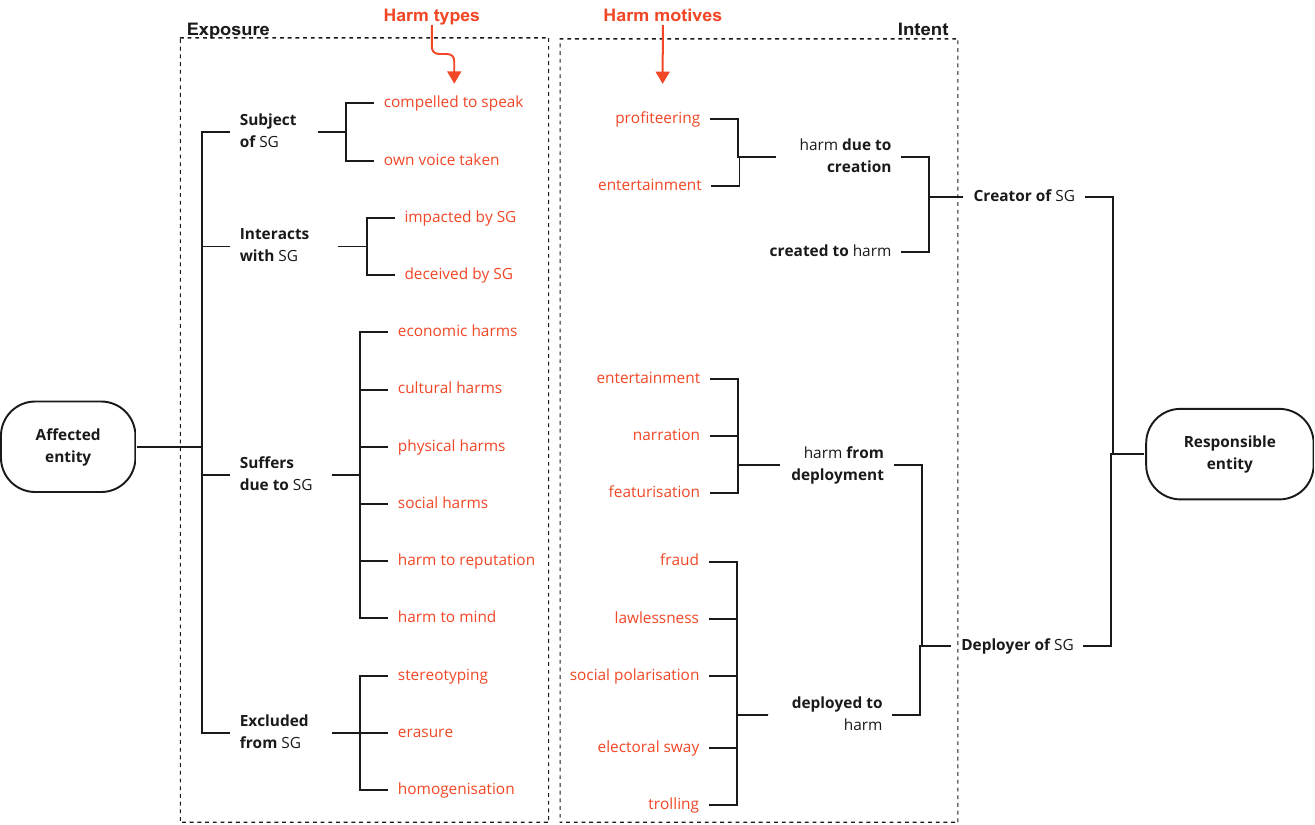}
    \caption{Taxonomy of harms of speech generators based on reported incidents. Harm types and motives in red expand the conceptual framework to the speech generation domain.}
    \label{fig:sg_harms_framework}
\end{figure}

\renewcommand*{\arraystretch}{1.35}
\begin{footnotesize}
\begin{longtable}{|L{0.15\textwidth}|L{0.28\textwidth}|L{0.3\textwidth}|L{0.2\textwidth}|} 
\caption{Theoretical extension of our taxonomy of harms to include specific harms from other AI harm taxonomies (\textcolor{blue}{in blue}).}\\ \hline
\label{tab:taxonomy_extension}
\textbf{Harm Type} & \textbf{Specific Harm} & \textbf{\citet{Shelby2022SociotechnicalReduction}} & \textbf{\citet{Weidinger2023SociotechnicalSystems}} \endhead \hline
\multicolumn{4}{c}{\textbf{Affected entity \textit{subject of} AI}} \\ \hline
Compelled to speak & Violation of right to self-determination & Loss of autonomy, loss of right to be forgotten & - \\
 & Violation of right to publicity & - & Violation of personal integrity \\ \hline
Own voice taken & Data laundering & Non-consensual data collection & Exploitative data sourcing and enrichment \\
 & Impersonation & - & Fraud \\ 
 & Identity theft & Privacy attack & Violation of personal integrity \\
 & Copyright theft & - & - \\
 & Identity hijack & - & - \\ 
 & Coercion & - & - \\ \hline
 \multicolumn{4}{c}{\textbf{Affected entity \textit{suffers due to} AI}} \\ \hline
Economic harms & Collusion & - & - \\
 & Loss of income & Economic loss & - \\
 & Loss of livelihood & Technological unemployment & - \\
 & Financial loss & \checkmark & - \\
 & \textit{\textcolor{blue}{Loss of opportunity}} & \textcolor{blue}{Opportunity loss} & \textcolor{blue}{Unfair distribution of benefits from model access} \\
 & \textit{\textcolor{blue}{Inadequate protection from financial loss}} & \textcolor{blue}{Discrimination in insurance, banking, or other financial sectors} & - \\ \hline
Cultural harms & Undermine creative economies & - & \checkmark \\
 & Cultural dispossession & Systemic erasure of culturally significant objects and practices & - \\
 & \textit{\textcolor{blue}{Misappropriation and exploitation}} & - & \textcolor{blue}{\checkmark} \\
 & \textit{\textcolor{blue}{Cultural hedgemony}} & \textcolor{blue}{\checkmark} & - \\
 & \textit{\textcolor{blue}{Proliferating false perceptions about cultural groups}} & \textcolor{blue}{\checkmark} & - \\ \hline
Physical harms & Bodily injury & \checkmark & - \\
 & Property damage & Inciting or enabling offline violence & - \\
 & \textit{\textcolor{blue}{Environmental damage}} & \textcolor{blue}{Environmental harms} & \textcolor{blue}{\checkmark} \\
 & \textit{\textcolor{blue}{Resource depletion}} & \textcolor{blue}{Depletion or contamination of natural resources} & - \\
 & \textit{\textcolor{blue}{Injury to animals}} & \textcolor{blue}{\checkmark} & - \\ \hline
Social harms & Abuse of public services & - & - \\
 & \textit{\textcolor{blue}{Systemic failure of critical systems}} & \textcolor{blue}{Systemic failures of financial systems} & - \\
 & Threat of terrorism & Nation destabilisation & - \\
 & National security threat & Nation destabilisation & Security threat \\
 & Inequality and precarity & Digital divides, labour exploitation & \checkmark \\
 & Pollution of information ecosystem & Malinformation, Misinformation & \checkmark \\
 & Erosion of trust in public information & - & \checkmark \\
 & Influence operations & Disinformation, Distortion of reality & \checkmark \\
 & \textit{\textcolor{blue}{Deteriorating social bonds}} & \textcolor{blue}{\checkmark} & - \\
 & \textit{\textcolor{blue}{Human rights violations}} & \textcolor{blue}{\checkmark} & - \\
 & \textit{\textcolor{blue}{Erosion of democracy}} & \textcolor{blue}{\checkmark} & - \\
 & \textit{\textcolor{blue}{Subjugating knowledges or foreclosing alternative ways of knowing}} & \textcolor{blue}{\checkmark} & - \\ \hline
Harm to reputation & Defamation & \multirow{2}{*}{Reputational harms} & \checkmark \\
 & Toxic content &  & \checkmark \\ \hline
Harm to mind & Deception & - & - \\ 
 & Mental disintegration & - & - \\
 & Emotional harms & \checkmark & - \\
 & Psychological distress & - & - \\
 & Online abuse & \checkmark & - \\
 & Privacy attack & \checkmark & Privacy infringement \\ \hline
\textit{\textcolor{blue}{Participation harms}} & \textit{\textcolor{blue}{Alienation}} & \textcolor{blue}{\checkmark} & - \\
 & \textit{\textcolor{blue}{Increased labour}} & \textcolor{blue}{\checkmark} & - \\
 & \textit{\textcolor{blue}{Service or benefit loss}} & \textcolor{blue}{\checkmark} & - \\
 & \textit{\textcolor{blue}{Loss of agency}} & \textcolor{blue}{\checkmark} &  \\ \hline
  \multicolumn{4}{c}{\textbf{Affected entity \textit{interacts with} AI}} \\ \hline
Impacted by & - & - & - \\
Deceived by & - & - & - \\
\textit{\textcolor{blue}{Reliant on}} & - & - & \textcolor{blue}{Overreliance} \\
\textit{\textcolor{blue}{Manipulated by}} & - & \textcolor{blue}{Behavioural manipulation} & \textcolor{blue}{Persuasion and manipulation} \\
\textit{\textcolor{blue}{Persuaded by}} & - & - & \textcolor{blue}{Persuasion and manipulation} \\
\textit{\textcolor{blue}{Dangerous information from}} & - & - & \textcolor{blue}{Dissemination of dangerous information} \\ 
\textit{\textcolor{blue}{False information from}} & - & - & \textcolor{blue}{Propagating misconceptions/ false beliefs} \\ \hline
\multicolumn{4}{c}{\textbf{Affected entity \textit{excluded from} AI}} \\ \hline
Stereotyping & - & \checkmark & \multirow{6}{*}{Unfair representation} \\ \cline{1-3}
Erasure & - & \checkmark &  \\ \cline{1-3}
Homogenisation & - & - &  \\ \cline{1-3}
\textit{\textcolor{blue}{Misrepresentation}} & \textit{\textcolor{blue}{Reifying essentialist social categories}} & \textcolor{blue}{\checkmark} &  \\
 & \textit{\textcolor{blue}{Denying people opportunity to self-identify}} & \textcolor{blue}{\checkmark} &  \\
 & \textit{\textcolor{blue}{Alienating social groups}} & \textcolor{blue}{\checkmark} &  \\ \hline
\end{longtable}
\end{footnotesize}

\end{document}